\documentclass{article}

\usepackage[preprint]{neurips_2025}

\usepackage[utf8]{inputenc} 
\usepackage[T1]{fontenc}    
\usepackage[colorlinks=true,citecolor=black,breaklinks=true,bookmarks=false,linkcolor=black]{hyperref} 
\usepackage{url}            
\usepackage{booktabs}       
\usepackage{xcolor}         
\usepackage{pifont}         

\usepackage{amsfonts}       
\usepackage{amsmath}        
\usepackage{nicefrac}       
\usepackage{array}          
\newcolumntype{C}[1]{>{\centering\arraybackslash}p{#1}}
\usepackage{microtype}      
\usepackage{graphicx}       
\usepackage{subcaption}     
\usepackage{verbatim}       
\usepackage{makecell}
\usepackage{pgfplots}
\pgfplotsset{compat=1.18}
\usepackage{caption}

\usepackage{algorithm}
\usepackage{algpseudocode}
\usepackage{algorithmicx}

\newcommand{\mycomment}[1]{\hfill $\triangleright$ #1}

\def \x {\mathbf{x}}
\def \xo {\mathbf{x}_{0}}
\def \xt {\mathbf{x}_{t}}
\def \sigt {\sigma(t)}
\def \sig {\sigma}

\def \N {\mathcal{N}}
\def \I {\mathbf{I}}
\def \o {\mathbf{0}}
\def \y {\mathbf{y}}
\def \w {\mathbf{w}}
\def \q {\mathbf{q}}
\def \ours {VideoPDE}

\def \u {\mathbf{u}}
\def \v {\mathbf{v}}
\def \m {\mathbf{m}}

\def \c {\boldsymbol{c}}
\def \t {\tau}

\def \y {\boldsymbol{y}}

\title{VideoPDE: Unified Generative PDE Solving \\via Video Inpainting Diffusion Models}

\author{
    Edward Li\thanks{Authors contribute equally to this work.}\quad Zichen Wang$^*$\quad Jiahe Huang$^*$\quad Jeong Joon Park\\
    University of Michigan, Ann Arbor\\
    \texttt{\{edwarli, zzzichen, chloehjh, jjparkcv\}@umich.edu}
}

\begin{document}

\maketitle


\vspace{-0.5cm}
\begin{abstract}
    \vspace{-0.3cm}
    We present a unified framework for solving partial differential equations (PDEs) using video-inpainting diffusion transformer models. Unlike existing methods that devise specialized strategies for either forward or inverse problems under full or partial observation, our approach unifies these tasks under a single, flexible generative framework. Specifically, we recast PDE-solving as a generalized inpainting problem, e.g., treating forward prediction as inferring missing spatiotemporal information of future states from initial conditions. To this end, we design a transformer-based architecture that conditions on arbitrary patterns of known data to infer missing values across time and space. Our method proposes pixel-space video diffusion models for fine-grained, high-fidelity inpainting and conditioning, while enhancing computational efficiency through hierarchical modeling. Extensive experiments show that our video inpainting-based diffusion model offers an accurate and versatile solution across a wide range of PDEs and problem setups, outperforming state-of-the-art baselines. Project page: \href{https://videopde.github.io/}{videopde.github.io}.
\end{abstract}


\vspace{-0.4cm}
\section{Introduction}
\vspace{-0.3cm}
Computational science communities have proposed numerous learning-based approaches for solving PDE-governed systems for simulation, optimization, or scientific discovery. These methods provide tradeoffs across accuracy, applicability, and speed. Physics-informed neural networks (PINNs) \cite{raissi2019physics,raissi2017physics} can be flexibly applied to forward/inverse predictions or sparse differential measurements, but their optimization often falls into a local minimum, sacrificing accuracy. Neural Operators \cite{li2020fourier,li2021physics,lu2021learning} offer fast approximate simulations, but struggle to handle partial observations common in real-world problems. Recent generative methods \cite{chen2023seine, Shu_2023, zhuang2024spatiallyawarediffusionmodelscrossattention} accommodate partial observations, at the expense of slow speed and inability to model dense temporal states. These challenges have limited the real-world applicability of learning-based PDE approaches for past state reconstructions or future simulations.

In this work, we introduce \ours{}, a unified framework that is accurate, fast, and applicable to a diverse range of scenarios. Intuitively, \ours{} casts the problem of PDE solving in diverse settings as a generative video inpainting problem, leveraging the power of diffusion models. For example, forward simulation can be viewed as predicting the missing pixels in frames 2 to T, given full or partial observations of the initial frame. This generative framework unifies diverse sensor configurations, supports non-deterministic predictions for chaotic systems, and offers a fast yet accurate alternative to classical solvers—all within a single neural network.

Our method uses a transformer-based video diffusion model (VDM) \cite{ho2022videodiffusionmodels} architecture that can be conditioned on arbitrary spatiotemporal patterns of partial measurements. Unlike common VDMs that operate in the latent space \cite{blattmann2023videoldm, gupta2023photorealisticvideogenerationdiffusion, he2022lvdm, zhou2023magicvideoefficientvideogeneration}, our method can denoise and condition at the pixel level, well-suited for scientific applications that require fine-grained accuracy, rather than perceptual realism. While VDMs have been used for video inpainting and prediction in the natural video domain \cite{hoppe2022diffusion, lu2023vdtgeneralpurposevideodiffusion, voleti2022MCVD, zhang2023avid}, no prior works (scientific or non-scientific domain) have tried hierarchical pixel-level modeling and intra-token conditioning that lead to exceptional accuracy and efficiency.

Our contributions include: (i) the formulation of casting PDE solving as generative video inpainting, (ii) architectural innovation that leads to efficient pixel-space denoising and conditioning, (iii) empirical demonstration of generative PDE solving across diverse settings. Our trained models obtain state-of-the-art results across settings, producing accurate predictions from as little as 1\% continuous measurements and reducing error by up to an order of magnitude compared to prior methods.


\section{Related Work}
\paragraph{Neural PDEs}

Solving partial differential equations (PDEs) is a fundamental problem in physical sciences. Traditional numerical methods such as the Finite Element Method \cite{quarteroni2008numerical, solin2005partial} and Boundary Element Method \cite{aliabadi2020boundary, idelsohn2003meshless} have long been the backbone of PDE solving but are computationally expensive and inflexible for complex systems. Neural approaches offer data-driven alternatives: physics-informed neural networks (PINNs) \cite{raissi2017physics, raissi2019physics} enforce PDE constraints via the loss function and have been applied to a wide range of PDEs \cite{cai2021physicsfluid, cai2021physicsheat, eivazi2022physics, hao2023pinnacle, mao2020physics, misyris2020physics, PENWARDEN2023111912, toscano2024pinnspikansrecentadvances, Waheed_2021}. While PINNs can work on sparse measurements, in practice they often face optimization instability and poor scalability. Neural operators, such as FNO \cite{li2020fourier}, DeepONet \cite{lu2021learning}, and PINO \cite{li2021physics}, learn mappings between function spaces to avoid expensive optimization and achieve resolution-invariance. These models have been extended to various forward \cite{bonev2023spherical, chetan2023accurate, li2023solving, li2022fourier, li2024geometry, peng2023linear, serrano2024operator, wen2022u} and inverse \cite{long2024invertible, molinaro2023neural} PDE tasks, but remain limited in flexibility for handling arbitrary and sparse input patterns.

\vspace{-0.2cm}
\paragraph{Solving PDEs Under Sparse Measurements}

Recently, neural methods have gained attention for solving PDEs under sparse measurements, reflecting the challenge of acquiring full spatiotemporal data. DiffusionPDE \cite{huang2024diffusionpdegenerativepdesolvingpartial} addresses this by modeling the joint distribution over coefficients and solutions, allowing flexible inputs, but its DPS \cite{chung2024diffusionposteriorsamplinggeneral} backbone requires PDE-specific tuning and struggles with dynamic PDEs. Spatially-aware diffusion models \cite{zhuang2024spatiallyawarediffusionmodelscrossattention} use cross-attention to handle partial observations but lack temporal modeling. Temporal PDEs are especially important for modeling nonlinear fluid and gas dynamics \cite{darman2025fourieranalysisphysicstransfer, dheeshjith2024samudraaiglobalocean, kashinath2021physics, wang2020towards, zong2023neural}. Super-resolution frameworks \cite{Fukami_2020, fukami2023super, Kim_2021} reconstruct full fields from coarse data. Recent methods \cite{li2023transformer, shan2024pirdphysicsinformedresidualdiffusion, Shu_2023} combine physics-informed losses with diffusion models or transformers for high-fidelity turbulent flow reconstruction.
Despite past successes, existing methods often rely on strong assumptions about PDEs, boundary conditions, or sensor layouts. We propose a unified generative framework that requires no prior knowledge and generalizes well across forward, inverse, and partial observation problems.

\vspace{-0.2cm}
\paragraph{Inpainting Diffusion Models}
Diffusion models \cite{ho2020denoising, sohldickstein2015deepunsupervisedlearningusing, song2020generativemodelingestimatinggradients, song2021scorebasedgenerativemodelingstochastic, song2022diffusionsurvey} have emerged as particularly suited for image and video inpainting due to their capability to model complex, high-dimensional distributions effectively. Training-free methods guide the sampling trajectory to satisfy the conditions at inference time through noise inference \cite{meng2022sdedit}, resampling \cite{lugmayr2022repaintinpaintingusingdenoising}, or latent alignment \cite{choi2021ilvrconditioningmethoddenoising}. It can also be studied as a linear inverse problem \cite{chung2024diffusionposteriorsamplinggeneral, chung2024improvingdiffusionmodelsinverse, kawar2022denoising, song2024solvinginverseproblemslatent, xu2023infedit}. However, these methods often struggle with extremely sparse or ambiguous observations. Another class of methods directly trains a conditional diffusion model. These methods typically modify the network architecture to inject conditioning information, such as channel concatenation \cite{saharia2022paletteimagetoimagediffusionmodels, saharia2021imagesuperresolutioniterativerefinement}, cross-attention \cite{balaji2023ediffitexttoimagediffusionmodels, perez2017filmvisualreasoninggeneral, rombach2022high}, or ControlNet \cite{zhang2023addingconditionalcontroltexttoimage}. We adopt channel concatenation in this work since it is simple and effective. These conditioning techniques have been extended to video diffusion models \cite{ho2022videodiffusionmodels} for video inpainting \cite{lu2024vdt, wang2023videocomposercompositionalvideosynthesis, zhang2024avidanylengthvideoinpainting}.


\begin{figure}
    \centering
    \includegraphics[width=\linewidth]{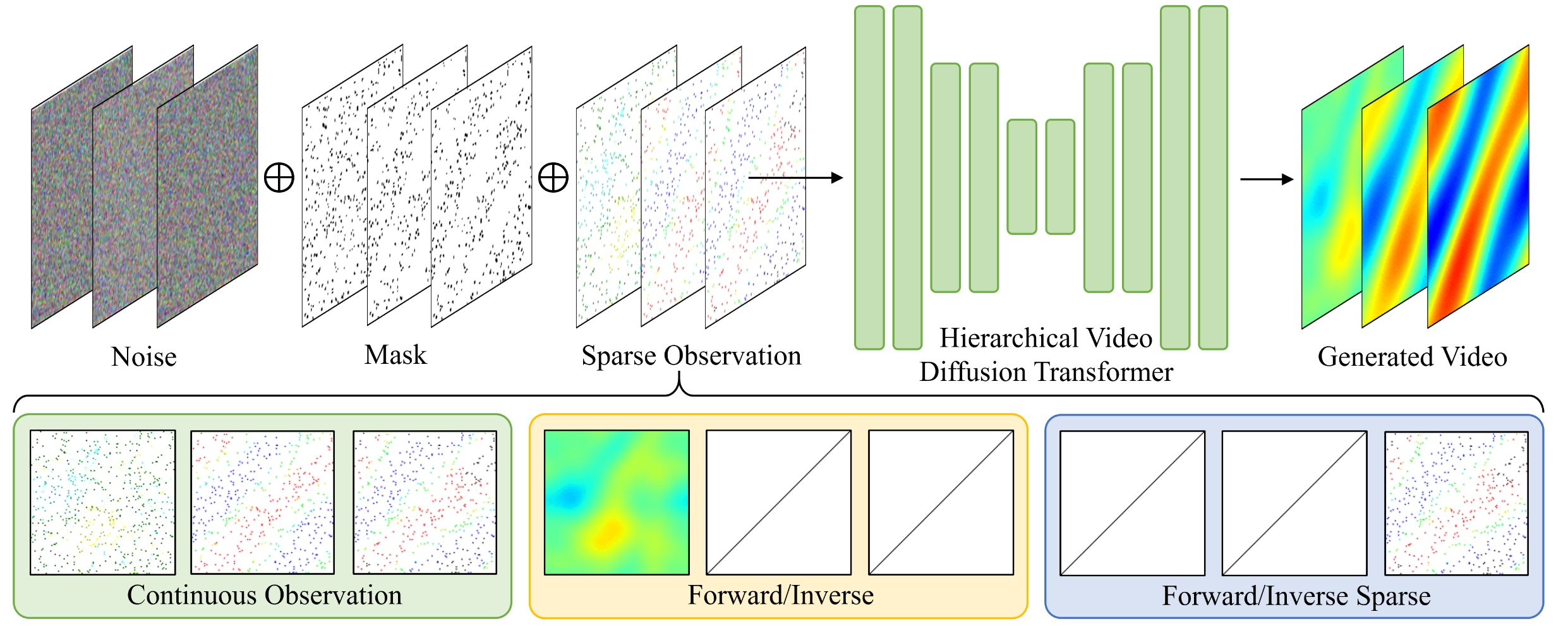}
    \caption{\textbf{\ours{} pipeline.} We cast PDE solving as a video inpainting task. Our Hierarchical Video Diffusion Transformer (HV-DiT) denoises initial noise into a full video, conditioned on pixel-level sparse measurements. Its ability to handle arbitrary input patterns enables flexible application to diverse PDE scenarios, including forward, inverse, and continuous measurement tasks.}
    \label{fig:architecture}
    \vspace{-0.4cm}
\end{figure}

\vspace{-0.2cm}
\section{Methods}
\vspace{-0.2cm}

\subsection{Preliminaries: Diffusion Models and Guided Sampling}

Diffusion models learn data distributions by reversing a gradual noising process. Starting from a clean sample $\xo$ from a data distribution $p(\x)$, a forward stochastic process progressively adds noise $\epsilon \sim \N(\o, \I)$ to produce $\xt = \xo + \sigt\epsilon$ and hence a family of distributions $p(\xt; \sig(t))$, where $\sig(t)$ denotes the standard deviation of the noise at diffusion time $t$, following the noise schedule of the Elucidating Diffusion Models (EDM) \cite{karras2022elucidating} framework we adopt in this work. The goal is to learn the \textit{reverse} process to recover $\xo$ from $\xt$ by training a denoising neural network $D_\theta(\xt, \sigt)$ with loss
\begin{align}
    \mathcal{L}_{\text{EDM}} = \mathbb{E}_{\xo \sim p(\x)} \mathbb{E}_{ \epsilon \sim \N(\o, \I)} \left[| D_\theta(\xt, \sigt) - \xo \|^2\right]
\end{align}
This gives us an estimate of the \textit{score function} \cite{song2020score}, a vector field pointing to higher data density,
\begin{equation}
    \nabla_{\x}\log p\big(\x;\sigt\big) = (D(\x,\sig(t))-\x) / \sig(t)^2,
\end{equation}
\vspace{-0.cm}
from which we can apply numerical ODE solvers to iteratively denoise from a complete noise $\x_{T} \sim \N(\o, \I)$ following
\vspace{-0.1cm}
\begin{equation}
    \mathrm{d}\x=-\dot{\sig}(t) \sigt \nabla_{\x}\log p\big(\x;\sigt\big) \mathrm{d}t.
\end{equation}
\vspace{-0.6cm}

\vspace{-0.2cm}
\paragraph{Conditional Sampling with Diffusion Models}
Diffusion models allow flexible input conditioning during inference sampling. I.e., conditional diffusion methods model a conditional data distribution $p(x|y; \sig(t))$ with a conditional score function $\nabla_{\x}\log p\big(\x|y)$. Conditional diffusion sampling can be roughly divided into methods that require computing gradients during inference and those that do not.

Gradient-based guidance \cite{dhariwal2021diffusion} approaches often maintain an unconditional network architecture, but use an auxiliary loss during inference to guide the score toward the conditioning. For example, DPS~\cite{chungdiffusion} approximates the conditional score function:
$    \nabla_{\x_t}\log p\big(\x_t|y)\approx \nabla_{\x_t}\log p\big(\x_t)-\zeta \nabla_{\x_t}\mathcal{L}_\phi(\x_t,y),
$
where $\mathcal{L}_\phi$ measures the current loss against the conditions $y$. While effective, gradient-based diffusion guidance methods tend to be slow and rely on intricate dynamics of the two directions, leading to hyperparameter sensitivity.

Gradient-free methods \cite{saharia2022paletteimagetoimagediffusionmodels,meng2021sdedit,ho2022classifier} typically train a specific network architecture that takes the conditioning $y$ as input. That is, the conditional denoising network models:
\begin{equation}
    \nabla_{\x}\log p\big(\xt|y;\sigt\big) \approx (D_\theta(\xt,y;\sigt)-\x) / \sigt^2.
\end{equation}
When there are significant ambiguities and noise associated with the conditioning $y$, e.g., when $y$ is in the form of text annotations, the network is incentivized to under-commit to the conditioning, leading to the classifier-free guidance (CFG)~\cite{ho2022classifier} technique to amplify the conditioning signal. In this work, we adopt the gradient-free network-conditioning strategy without using CFG.

\subsection{Spatiotemporal PDE Solving as Video Inpainting}
\begin{figure}
    \vspace{0.5em}
    \centering
    \includegraphics[width=\linewidth]{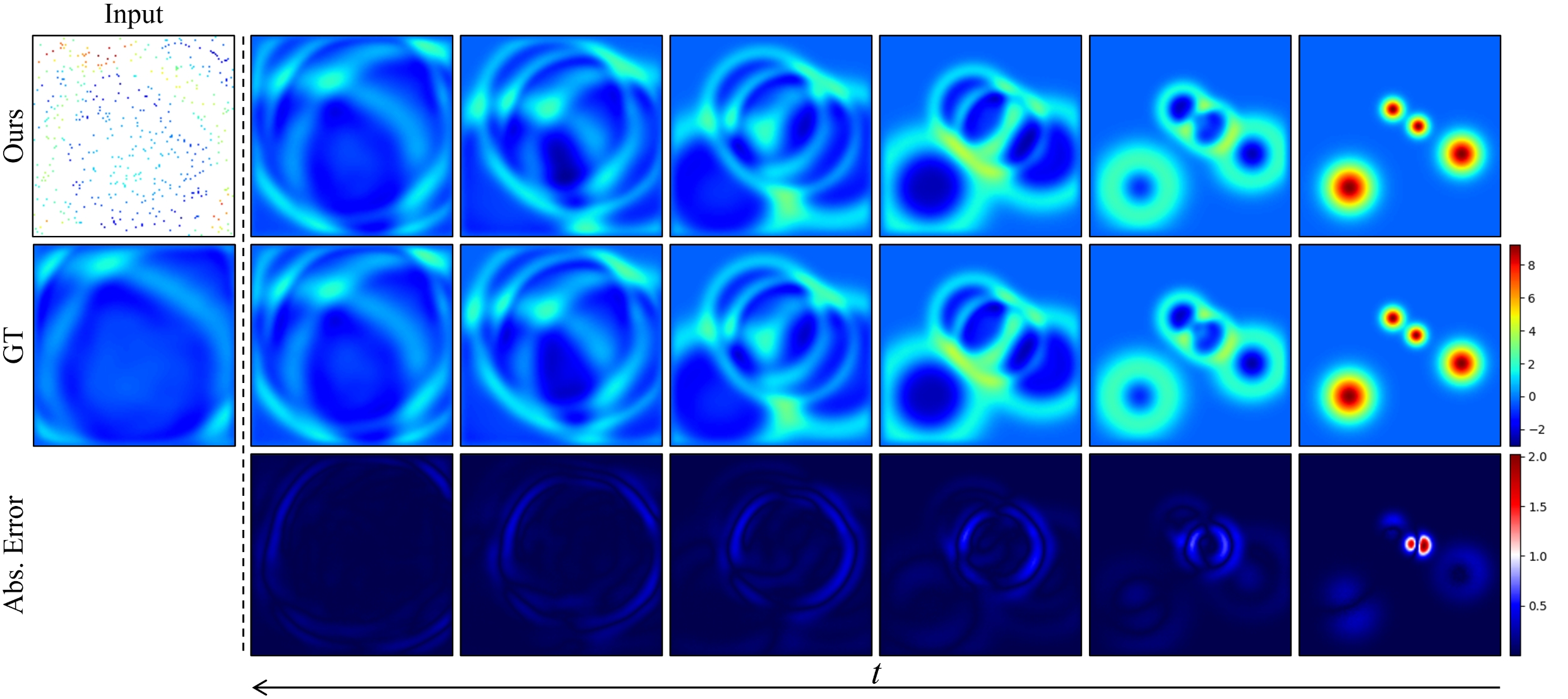}
    \caption{\textbf{Inverse simulation from partial observation.} \ours{} formulates general PDE solving as a video inpainting problem, where unknown pixels are denoised conditioned on sparse inputs. Here, given 3\% observation at time \(T\), \ours{} accurately recovers the whole trajectory \(T-1\to1\).}
    \label{fig:inverse}
    \vspace{-0.4cm}
\end{figure}

We cast the problem of spatiotemporal PDE solving as a video inpainting task, enabling a unified and flexible framework for handling a wide range of prediction scenarios (\autoref{fig:inverse}). Like prior data-driven PDE approaches, our goal is to learn a neural network that can infer unknown system states across a family of equations. However, unlike existing methods that typically design separate models for forward, inverse, or partially observed cases, our approach treats all such tasks as instances of conditional video inpainting.

In this formulation, we cast PDE solving as the task of filling in missing regions of a video representing the evolution of physical states over time and space. For example, forward prediction corresponds to inpainting future frames based on an initial condition; partially observed setups correspond to inpainting from sparse spatiotemporal sensor data. Our proposed architecture, described in detail in Section \ref{sec:architecture}, is a transformer-based diffusion model explicitly designed to condition on arbitrary patterns of observed data and generate coherent, accurate completions.

\vspace{-0.2cm}
\paragraph{PDE Formulation}
While our formulation accommodates both static (time-independent) and dynamic (time-dependent) PDEs, we focus on dynamic systems, e.g., Navier–Stokes:
\begin{equation}\label{eq:dynamic}
    \begin{alignedat}{2}
        f(\c,\t; \u) & = 0, \qquad                     &  & \text{in } \Omega \times (0,\infty),                    \\
        \u(\c,\t)    & = \boldsymbol{g}(\c,\t), \qquad &  & \text{on } \partial \Omega \times (0,\infty),           \\
        \u(\c,\t)    & = \boldsymbol{o}(\c,\t), \qquad &  & \text{on } \mathcal{O} \subset \Omega \times (0,\infty)
    \end{alignedat}
\end{equation}
Here, $\c$ and $\t$ denote the spatial and temporal coordinates, respectively, and $\u(\c,\t)$ is the solution field. The boundary condition is given by $\u|_{\partial \Omega \times (0,\infty)} = \boldsymbol{g}$. We aim to recover the full solution $\u_\t$ at any time $\t \in [0, T]$ from sparse spatiotemporal observations $\mathcal{O}$, where $\u|_{\mathcal{O}} = \boldsymbol{o}$. We make no assumptions about the structure of these observed locations.

\vspace{-0.2cm}
\paragraph{Diffusion-based Video Inpainting}
We cast PDE-solving as a spatiotemporal inpainting task, where missing regions of the solution field $\u(\c, \t)$ are inferred from sparse observations $\mathcal{O}$. To solve this inpainting problem, we leverage the powerful generative capabilities of diffusion models. Specifically, we train a conditional diffusion model to learn the distribution of physically consistent video-like solution trajectories, while conditioning on arbitrary known subsets of the spatiotemporal domain.

We represent each PDE solution as a video $\x \in \mathbb{R}^{H \times W \times T \times C}$, where $H \times W$ is the spatial grid, $T$ is the number of time steps, and $C$ the number of field channels. The conditioning signal is defined by a binary mask $\m \in \{0,1\}^{H \times W \times T}$ and corresponding observed values $\y = \x \odot \m$. During training, we sample random spatiotemporal masks and supervise the model to reconstruct the full video from these partial views. The model learns the conditional score function:
\begin{equation}
    \nabla_{\x}\log p(\x|\y;\sigt) \approx ({D_\theta(\x_t, \y, \m; \sigt) - \x_t})/\sigt^2,
\end{equation}
where $D_\theta$ is a transformer-based denoising network conditioned on $\y$ and $\m$, and $\x_t$ is a noisy intermediate sample at diffusion time $t$. During inference, we take sparse observations $\y$ and $\m$ as inputs, initialize $\x_T$ with pure Gaussian noise, and denoise it using the learned score function.

By casting PDE-solving as conditional video generation, we unify a broad class of spatiotemporal problems under a generative modeling task. Importantly, our formulation enables conditioning the same model on different observation patterns, e.g. forward and inverse predictions, or interpolation from arbitrary observed subsets. Section~\ref{sec:architecture} details the model design and training process.

\subsection{Hierarchical Video Diffusion Transformer (HV-DiT)}\label{sec:architecture}

While most recent state-of-the-art diffusion models \cite{rombach2022high} operate in a learned latent space to reduce computational cost, we design our architecture to perform diffusion directly in pixel space, as shown in Figure \ref{fig:architecture}. This choice is motivated by our observation that pixel-space diffusion yields significantly more accurate and physically consistent reconstructions, which is particularly important in PDE settings where fine-grained field values matter more than perceptual qualities.

To manage the high dimensionality of pixel-space video data, we tokenize each input video $\x \in \mathbb{R}^{H \times W \times T \times C}$ by merging small spatiotemporal neighborhoods, for example, $N \times N \times N$ patches, into single tokens. This results in a structured token sequence over which we design an efficient variant of the Video DiT architecture \cite{openai2024sora}, which we refer to as \textbf{HV-DiT}, inspired by the hierarchical image model HDiT \cite{crowson2024scalable}. Unlike standard transformers with global self-attention, HV-DiT employs localized attention, restricting each token's receptive field to nearby spatiotemporal neighbors. This reduces computational complexity and allows the model to focus on local PDE dynamics.

Our transformer architecture is hierarchical \cite{crowson2024scalable, nawrot2021hierarchical}: tokens are progressively downsampled by merging neighboring tokens, creating a multi-scale representation. This downsampling path is paired with an upsampling path with skip connections in the style of U-Net, enabling both local detail preservation and global context integration. At each layer, we apply spatiotemporal neighborhood attention. At the coarsest resolution (bottleneck), we use global attention layers to capture long-range spatiotemporal dependencies.

A key architectural innovation is the way we condition the model on known observations. For each token, we concatenate its associated binary mask (indicating observed pixels) and the corresponding observed values. This allows our model to condition at the individual pixel level, enabling fine-grained, spatially varying guidance during the denoising process. Concatenating the binary mask resolves ambiguity between observed and unobserved pixels. This formulation supports flexible conditioning across a wide range of scenarios, including forward prediction, inverse recovery, and inpainting from arbitrary subsets of observations. The concatenated final input to $D_\theta(\x_t^{cond})$ is:
\vspace{-0.1cm}
\begin{equation}
    \x_t^{cond}\equiv \text{concat}(\x_t, \m, \y), \quad \text{\# of tokens is } H/N\times W/N\times T/N
\end{equation}
\vspace{-0.1cm}
Note that only the solution field $\x$ part of the input token contains the diffusion noise.

Overall, our HV-DiT combines the expressiveness of pixel-space modeling with the efficiency of localized and hierarchical attention, forming a powerful and versatile backbone for generative PDE-solving through conditional video inpainting.


\begin{table}[t]
    \centering
    \setlength{\tabcolsep}{3.pt}
    \begin{tabular}{
            l  
            l  
            c  
            c  
            c  
            c  
            c  
            c  
        }
        \toprule
        \textbf{Method}                                                            &
        \textbf{Type}                                                              &
        \makecell{\textbf{Partial}                                                                                                                                           \\ \textbf{obs.}} &
        \makecell{\textbf{Flexible}                                                                                                                                          \\ \textbf{inference}} &
        \makecell{\textbf{Dense}                                                                                                                                             \\ \textbf{temporal}} &
        \makecell{\textbf{Inf.}                                                                                                                                              \\ \textbf{Time (s)}$\downarrow$} &
        \makecell{\textbf{Forward}                                                                                                                                           \\ \textbf{Error}$\downarrow$} \\
        \midrule
        PINN \cite{raissi2019physics}                                              & PDE loss            & \checkmark & \checkmark  & \checkmark & 66   & 27.3\%             \\
        FNO \cite{li2020fourier}                                                   & Neural Operator     &            &             & \checkmark & 2.0  & \underline{2.7\%}  \\
        DeepONet \cite{lu2021learning}                                             & Neural Operator     &            &             & \checkmark & 1.7  & 11.3\%             \\
        PINO \cite{li2021physics}                                                  & Neural Operator+PDE &            &             & \checkmark & 2.2  & 4.9\%              \\
        DiffusionPDE \cite{huang2024diffusionpdegenerativepdesolvingpartial}       & Generative+PDE      & \checkmark & $\triangle$ &            & 4480 & 5.8\%              \\
        Shu et al. \cite{Shu_2023}                                                 & Generative+PDE      & \checkmark &             &            & 5760 & \text{\sffamily X} \\
        Zhuang et al. \cite{zhuang2024spatiallyawarediffusionmodelscrossattention} & Generative          & \checkmark &             &            & 4860 & \text{\sffamily X} \\
        \ours{} (Ours)                                                             & Generative          & \checkmark & \checkmark  & \checkmark & 9.3  & \textbf{0.45\%}    \\
        \bottomrule
    \end{tabular}
    \vspace{0.1cm}
    \caption{
        \textbf{Conceptual comparison of PDE-solving methods.} Neural operator methods struggle with partial inputs. Only PINN and \ours{} handle forward, inverse, and continuous measurements flexibly. Generative baselines focus on reconstructing one or two frames (instead of dense temporal frames) and are often not designed for forward prediction, where \ours{} excels.
    }
    \label{tab:comparison}
    \vspace{-0.6cm}
\end{table}

\vspace{-0.3cm}
\section{Experiments}
\vspace{-0.3cm}
We comprehensively evaluate \ours{}'s ability to solve a range of temporal PDEs across diverse inference scenarios. Specifically, we assess its performance in (i) reconstructing from continuous spatiotemporal sensor measurements (Table~\ref{tab:continuous_partial}), (ii) predicting future or past system states (Table~\ref{tab:forward_inverse_full}), (iii) handling partial observations during forward and inverse prediction (Table~\ref{tab:forward_inverse_partial}) , and (iv) generalizing across multiple inference tasks, including forward, inverse, and reconstruction.

\vspace{-0.2cm}
\paragraph{Baselines}
We compare \ours{} against a representative set of learning-based PDE solvers. For standard forward and inverse prediction under full initial or final conditions, we include FNO, PINO, DeepONet, and DiffusionPDE, each representing a distinct modeling paradigm (see Table~\ref{tab:comparison}). For partial observation settings, we compare only against DiffusionPDE, which has demonstrated superior performance and shown that prior baselines struggle with sparse conditioning. For the continuous measurement reconstruction task, we evaluate against state-of-the-art generative methods, including those proposed by Shu et al.\ \cite{Shu_2023}, Zhuang et al.\ \cite{zhuang2024spatiallyawarediffusionmodelscrossattention}, and DiffusionPDE \cite{huang2024diffusionpdegenerativepdesolvingpartial}. We also extend DiffusionPDE for improved temporal message passing. See the supplementary for more details.

\vspace{-0.2cm}
\subsection{PDE Problem Settings}\label{sec:problem-settings}
\vspace{-0.15cm}
We show the effectiveness of \ours{} mainly on 2D dynamic PDEs for forward, inverse, and continuous observation problems and compare it against SOTA learning-based techniques. We use the following families of PDEs for the main experiments. We refer readers to the supplementary for a more comprehensive coverage of our experiments

\vspace{-0.25cm}
\paragraph{Wave-Layer}
We evaluate our method on the Wave-Layer task following Poseidon \cite{herde2024poseidon}.
This task is based on the wave equation with spatially varying propagation speed and absorbing boundary:
\[\partial_t^2\u(\c,\tau)+(\q(\c))^2\Delta\u(\c,\tau)=0,\quad(\c,\tau)\in\Omega\times(0,T).\]
Here, \(\u\colon\Omega\times(0,T)\to\mathbb{R}\) is a scalar field representing displacement, and \(\q\colon\Omega\to\mathbb{R}\) represents propagation speed.
The initial condition is the sum of 2-6 Gaussians with random location and scale:
The propagation speed coefficient \(\c\) is generated by creating 3-6 layers with piecewise constant propagation speeds.
The layers are separated by reandomly generated frontiers.
The dataset contains 10,512 trajectories, each with 21 time steps at \(128\times128\) resolution.
The final 100 trajectories are used for validation and the rest for training.
This task arises from propagation of seismic waves through a layered medium. See the supplementary for more details on this problem.

\vspace{-0.25cm}
\paragraph{Navier–Stokes Equation}
We study the two-dimensional incompressible Navier–Stokes equations in vorticity form, following the setup introduced in DiffusionPDE~\cite{huang2024diffusionpdegenerativepdesolvingpartial}:
\begin{equation}
    \begin{aligned}
        \partial_t\w(\c, \t) + \v(\c, \t)\cdot\nabla \w(\c, \t) & = \nu\Delta \w(\c, \t) + \q(\c), &  & \c\in\Omega,\, \t\in(0,T], \\
        \nabla\cdot \v(\c, \t)                                  & = 0,                             &  & \c\in\Omega,\, \t\in[0,T], \\
        \w(\c, 0)                                               & = \w_0(\c),                      &  & \c\in\Omega.
    \end{aligned}
\end{equation}
Here, $\w = \nabla \times \v$ denotes the vorticity field, and $\v(\c, \t)$ is the velocity field at spatial location $\c$ and time $\t$. We fix the viscosity coefficient to $\nu = 10^{-3}$, corresponding to a Reynolds number of $Re = 1/\nu = 1000$. Initial conditions $\w_0$ are sampled from a Gaussian random field as in DiffusionPDE. Each datapoint is composed of 20 frames of a $128\times128$ vorticity field $\w$.

The external forcing $\q(\c)$ determines the long-term behavior of the system. In this setting, we adopt a static, time-independent forcing term:
\[
    \q(\c) = 0.1\left(\sin(2\pi(c_1 + c_2)) + \cos(2\pi(c_1 + c_2))\right),
\]
which introduces smooth, low-frequency energy into the system without any feedback from the flow itself. Due to the weak magnitude of this forcing and the absence of dynamic coupling, the system exhibits diffusion-like decay: initial high-frequency vorticity structures dissipate over time as the system evolves under viscous damping.

\vspace{-0.25cm}
\paragraph{Kolmogorov Flow}
To study more complex and persistent flow dynamics, we also evaluate our method on the Kolmogorov flow (KF) \cite{chandler2013invariant}, a classical setup used in \cite{Shu_2023} to simulate forced, quasi-turbulent regimes in 2D fluid dynamics. The same Navier–Stokes formulation applies, but with a different forcing term of the KF form:
\[
    \q(\c, \t) = -4\cos(4c_2) - 0.1\w(\c, \t).
\]
This forcing is composed of a strong, anisotropic spatial component ($\cos(4c_2)$) that continuously injects energy into the system, and a linear drag term ($-0.1\w$) that stabilizes the flow by removing energy at small scales. Crucially, the forcing depends on the evolving state $\w(\c, \t)$, introducing dynamic feedback that enables sustained motion.

Unlike the decaying dynamics of the previous setup, Kolmogorov flow exhibits persistent, swirling structures and high-frequency vorticity patterns over time. This makes it a challenging and realistic benchmark for generative PDE modeling, particularly in capturing long-term, high-fidelity spatiotemporal behavior. Finally, each datapoint is a 20-frame $256\times256$ vorticity field.

\vspace{-0.2cm}
\subsection{Experiment Details} \label{sec:experiment-details}
\vspace{-0.2cm}
We provide additional details on the datasets and their processing in the supplementary.
Training is performed on 4 NVIDIA L40S GPUs with a batch size of 8 per GPU, taking approximately 24 hours per model. All models are trained until convergence using the Adam optimizer with a constant learning rate schedule (initial LR $5 \times 10^{-4}$).
Our HV-DiT architecture operates directly in pixel space. Videos are tokenized into $4 \times 4 \times 2$ patches, forming a $32 \times 32 \times 10$ token grid with embedding dimension \(384\) for WL and NS. The model uses 2 transformer layers with neighborhood attention (window size $7 \times 7 \times 2$) and a downsampling operation via patch merging with factor \(2\). We provide more details on the model architecture and hyperparameters in the supplementary.

\vspace{-0.2cm}
\subsection{Experiment Results}\label{sec:experiment-results}
\vspace{-0.1cm}
\begin{table*}[t]
    \centering
    \begin{tabular}{l cc cc cc}
        \toprule
                            & \multicolumn{2}{c}{\textbf{Wave-Layer}} & \multicolumn{2}{c}{\textbf{Navier–Stokes}} & \multicolumn{2}{c}{\textbf{Kolmogorov Flow}}                                                                \\
                            & 1\%                                     & 3\%                                        & 1\%                                          & 3\%                & 1\%                & 3\%                \\
        \midrule
        DiffusionPDE        & 48.3\%                                  & 17.4\%                                     & 4.7\%                                        & 3.7\%              & 20.3\%             & 11.9\%             \\
        DiffusionPDE (Ext.) & 45.2\%                                  & 15.6\%                                     & 4.1\%                                        & 3.4\%              & 19.5\%             & 10.3\%             \\
        Shu et al.          & 49.7\%                                  & 17.7\%                                     & 8.6\%                                        & 6.2\%              & 19.7\%             & 11.8\%             \\
        Zhuang et al.       & 29.9\%                                  & 10.3\%                                     & 12.7\%                                       & 4.8\%              & 13.9\%             & 6.1\%              \\
        \midrule
        Ours                & ~\textbf{2.62\%}                        & ~\textbf{1.57\%}                           & \textbf{0.80\%}                              & \textbf{0.44\%}    & \textbf{6.48\%}    & \underline{2.71\%} \\
        Ours (unified)      & \underline{3.97\%}                      & \underline{2.05\%}                         & \underline{1.13\%}                           & \underline{0.48\%} & \underline{7.59\%} & \textbf{2.55\%}    \\
        \bottomrule
    \end{tabular}
    \caption{\textbf{Continuous partial observation reconstruction.} We quantitatively measure the performance of different methods using average \(\ell_2\) relative errors on Wave-Layer, Navier–Stokes, and Kolmogorov Flow benchmarks with 1\% and 3\% observation points.}
    \label{tab:continuous_partial}
    \vspace{-0.3cm}
\end{table*}

\begin{figure}
    \vspace{0.2em}
    \centering
    \includegraphics[width=\linewidth]{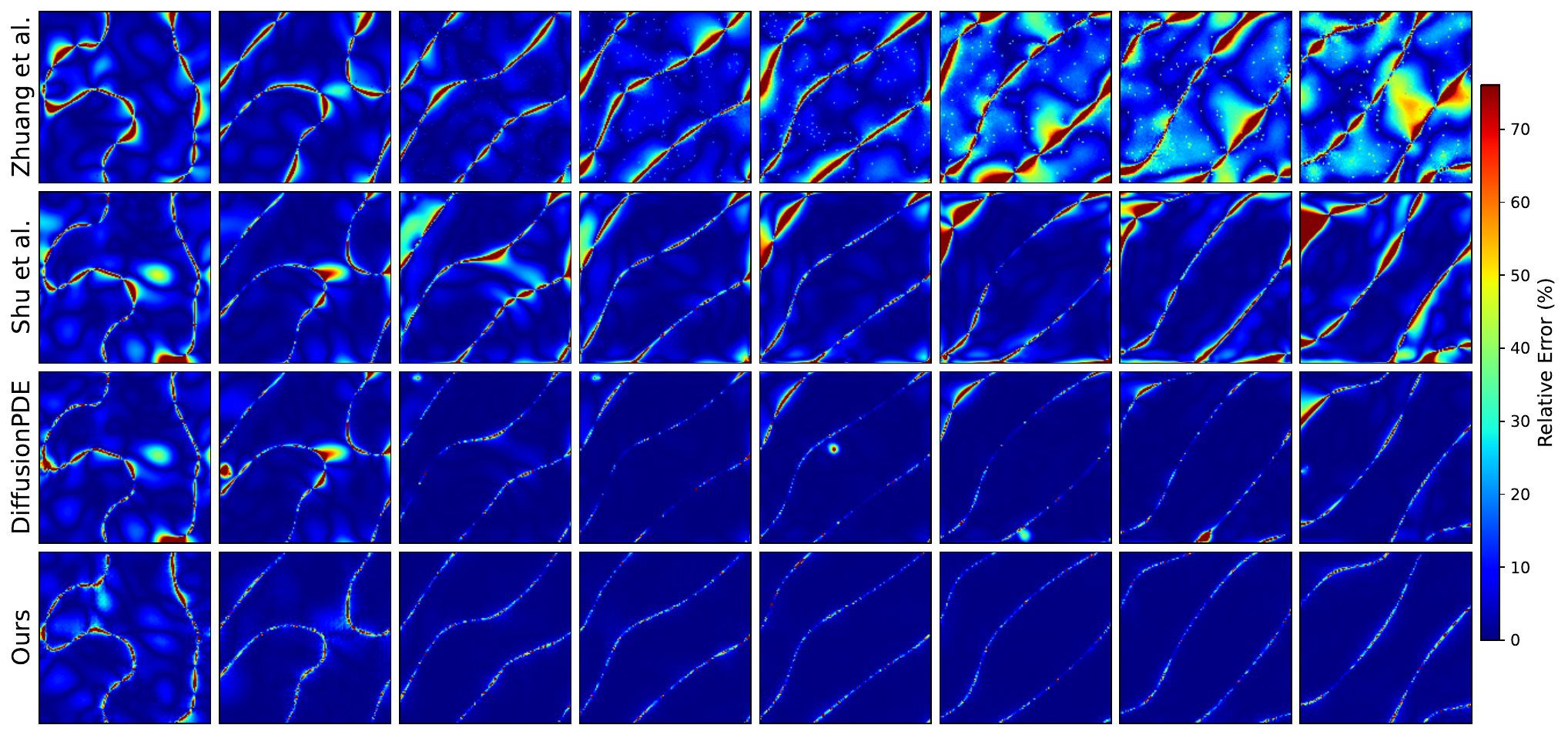}
    \caption{\textbf{Continuous measurement reconstruction comparison.}
        We compare relative error maps for reconstructing dense spatiotemporal fields from fixed sensors providing 1\% continuous observations on Navier–Stokes. Our results are the most accurate with minimal error. In contrast, baseline methods are significantly slower and not suitable for forward prediction (Zhuang \& Shu).
    }
    \label{fig:continuous}
    \vspace{-0.5cm}
\end{figure}

\paragraph{Continuous Partial Observation}
We evaluate the ability of \ours{} to reconstruct full spatiotemporal PDE trajectories from sparse, fixed-point observations. Specifically, we randomly sample a very small percentage of spatial coordinates (1\% or 3\%) and provide the solution values across all time steps at those locations. This setting mimics real-world sensor deployments, where measurements are collected continuously at fixed spatial positions. Our model is conditioned on these sparse yet temporally continuous observations.

As shown in Table~\ref{tab:continuous_partial}, we report the relative $\ell_2$ error across 100 held-out trajectories for three PDEs: Wave-Layer, Navier–Stokes, and Komolgorov Flow. In Figure \ref{fig:continuous} we visualize the error map for the Navier–Stokes equation. Our method significantly outperforms existing generative baselines, including DiffusionPDE \cite{huang2024diffusionpdegenerativepdesolvingpartial}, Shu et al. \cite{Shu_2023}, and Zhuang et al. \cite{zhuang2024spatiallyawarediffusionmodelscrossattention}, up to an order of magnitude, demonstrating robust generalization under extreme observation sparsity.

\begin{table*}[t]
    \centering
    \begin{tabular}{l cc cc cc}
        \toprule
                       & \multicolumn{2}{c}{\textbf{Wave-Layer}} & \multicolumn{2}{c}{\textbf{Navier–Stokes}} & \multicolumn{2}{c}{\textbf{Kolmogorov Flow}}                                                                \\
                       & Forward                                 & Inverse                                    & Forward                                      & Inverse            & Forward            & Inverse            \\
        \midrule
        FNO            & 35.34\%                                 & 65.43\%                                    & 2.71\%                                       & \underline{7.62}\% & 56.43\%            & 59.42\%            \\
        PINO           & 10.8\%                                  & 19.7\%                                     & 4.9\%                                        & \textbf{6.9\%}     & 7.4\%              & 7.6\%              \\
        DeepONet       & 47.68\%                                 & 53.32\%                                    & 11.29\%                                      & 12.63\%            & 46.61\%            & 46.92\%            \\
        DiffusionPDE   & 6.7\%                                   & 14.2\%                                     & 6.1\%                                        & 8.6\%              & 9.1\%              & 10.8\%             \\
        \midrule
        Ours           & ~\textbf{1.21\%}                        & ~\textbf{5.24\%}                           & \textbf{0.45\%}                              & 9.87\%             & \textbf{2.95\%}    & \textbf{4.90\%}    \\
        Ours (unified) & \underline{1.53\%}                      & \underline{12.65\%}                        & \underline{1.52\%}                           & 10.37\%            & \underline{4.60\%} & \underline{4.99\%} \\
        \bottomrule
    \end{tabular}
    \caption{\textbf{Forward/inverse full observation.} Average \(\ell_2\) relative errors of baseline methods for forward and inverse subtasks across datasets.}
    \vspace{-0.5cm}
    \label{tab:forward_inverse_full}
\end{table*}
\vspace{-0.2cm}

\begin{figure}
    \centering
    \includegraphics[width=\linewidth]{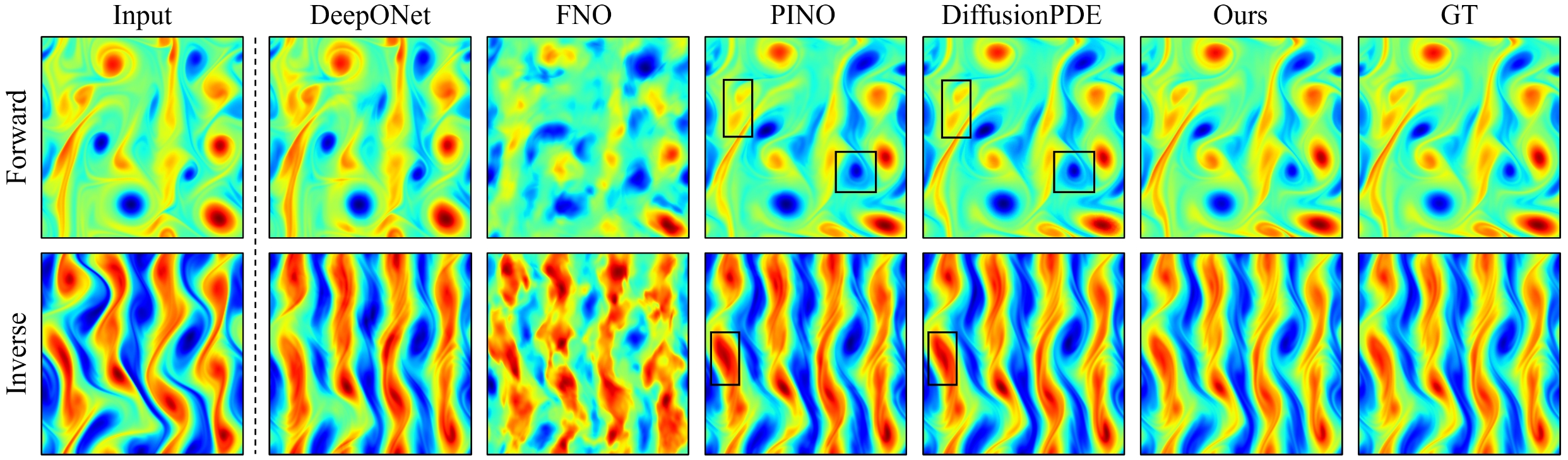}
    \caption{\textbf{Comparison of forward/inverse predictions on Kolmogorov Flow.} Our predictions are perceptually indistinguishable from ground truths, whereas other baseline results exhibit deviations. Significant errors of PINO and DiffusionPDE are squared in black.}
    \label{fig:foward-inverse-full}
    \vspace{-0.3cm}
\end{figure}

\paragraph{Forward/Inverse Full Observation}
We evaluate \ours{} on reconstructing full PDE trajectories given a single frame at either the start (forward prediction) or end (inverse inference) of the sequence. The full conditioning frame is provided while the remaining frames are masked. This setup reflects practical simulation scenarios where dense initial conditions are available and parallels image-to-video tasks in generative modeling.

Figure \ref{fig:foward-inverse-full} shows the final/initial frames of the fully observed forward/inverse processes on the Kolmogorov Flow dataset, demonstrating that \ours{} consistently produces results that are closer to the ground truth.
Table~\ref{tab:forward_inverse_full} reports the relative $\ell_2$ error across 100 held-out trajectories for three PDEs. \ours{} consistently outperforms baselines in both forward and inverse tasks, except for the low-frequency inverse setting. We attribute this to aleatoric uncertainty: in the NS dataset, diffusive dynamics lead to low-frequency end states that may originate from many high-frequency initial conditions. In such cases, pixel-wise $\ell_2$ loss penalizes plausible reconstructions and favors blurry averages. We leave exploration of distribution-based evaluation metrics for future work.

\vspace{-0.2cm}
\paragraph{Forward/Inverse Partial Observation}
We extend the forward and inverse prediction tasks to the partially observed setting by conditioning on a single frame—either at the start or end of the trajectory, with only 3\% of spatial points revealed. The model must reconstruct the full trajectory from these observations, reflecting real-world scenarios where sensors provide limited data at a single timepoint.

In Figure \ref{fig:inverse}, we present the inverse simulation of a Wave-Layer sample, where \ours{} recovers all time steps in reverse given only 3\% of observation points from the final frame. As shown in Table~\ref{tab:forward_inverse_partial}, \ours{} outperforms DiffusionPDE, the current SOTA for this task, by a significant margin across all settings, except for inverse prediction on the Navier–Stokes case, where aleatoric uncertainty remains high due to the diffusive loss of high-frequency information.
We note that \ours{} performs similarly on this task to the forward/inverse \emph{full} observation task, particularly for Wave-Layer forward prediction, and both Navier–Stokes forward and inverse prediction.

\begin{table*}[t]
    \centering
    \begin{tabular}{l cc cc cc}
        \toprule
                       & \multicolumn{2}{c}{\textbf{Wave-Layer}} & \multicolumn{2}{c}{\textbf{Navier–Stokes}} & \multicolumn{2}{c}{\textbf{Kolmogorov Flow}}                                                                   \\
                       & Forward                                 & Inverse                                    & Forward                                      & Inverse             & Forward             & Inverse             \\
        \midrule
        DiffusionPDE   & 19.5\%                                  & 24.3\%                                     & 3.9\%                                        & \textbf{10.2\%}     & 28.2\%              & 32.6\%              \\
        \midrule
        Ours           & ~\textbf{1.40\%}                        & ~\textbf{11.81\%}                          & \textbf{0.71\%}                              & \underline{10.41\%} & \textbf{11.66\%}    & \textbf{13.50\%}    \\
        Ours (unified) & \underline{1.89\%}                      & \underline{18.56\%}                        & \underline{1.61\%}                           & 11.45\%             & \underline{16.11\%} & \underline{25.75\%} \\
        \bottomrule
    \end{tabular}
    \caption{\textbf{Forward/inverse 3\% observation.} Average \(\ell_2\) relative errors of baseline methods for forward and inverse subtasks across datasets.}
    \label{tab:forward_inverse_partial}
    \vspace{-0.5cm}
\end{table*}

\begin{table*}[t]
    \vspace{0.7em}
    \centering
    \begin{tabular}{lr}
        \toprule
                                                                & Navier–Stokes      \\
        \midrule
        DiT with mixed noise and conditioning tokens            & 50.77\%            \\
        DiT with each token concatenated on sparse observations & 1.46\%             \\
        + Latent diffusion                                      & 7.13\%             \\
        + 3D Neighborhood attention and downsampling            & \underline{0.73\%} \\
        \midrule
        + concatenating binary mask (Ours, HV-DiT)              & \textbf{0.44\%}    \\
        \bottomrule
    \end{tabular}
    \caption{\textbf{Ablation study.} We ablate our design choices, beginning with a latent space DiT. We report average relative \(\ell_2\) errors for all configurations on Navier–Stokes with 3\% observation rate.}
    \label{tab:ablation}
    \vspace{-0.3cm}
\end{table*}

\vspace{-0.2cm}
\paragraph{Unified Model}
We evaluate whether a single model can jointly learn multiple inference tasks within our video inpainting framework. For each dataset, we train one unified model on six tasks: continuous partial observation (3\% and 1\%), forward and inverse prediction under full/partial observation.

As shown in Tables~\ref{tab:continuous_partial}, \ref{tab:forward_inverse_full}, and \ref{tab:forward_inverse_partial}, the unified model matches the performance of task-specific variants and outperforms prior baselines in most settings. In contrast, all baselines require separate models per task, highlighting \ours{}'s potential to be a unified framework for flexible PDE solving.

\vspace{-0.3cm}
\subsection{Ablation Study}
\vspace{-0.2cm}
We conduct an ablation study to assess the impact of key architectural choices, evaluated on the continuous partial observation task for low-frequency Navier–Stokes with a 3\% observation rate. Relative $\ell_2$ errors are reported in Table~\ref{tab:ablation}.

We begin with a video DiT architecture adapted from VDT \cite{lu2024vdt}, originally designed for natural video inpainting. The model input is \(\mathbf{x}\odot(1-\mathbf{m})+\mathbf{y}\odot\mathbf{m}\), where \(\mathbf{x}\) is Gaussian noise, \(\mathbf{m}\) a binary mask, and \(\mathbf{y}\) the ground truth. This model performed poorly, likely due to confusion between sparse observations and noise.

Replacing the conditioning method with channel-wise concatenation of noise and masked ground truth, i.e., \(\mathrm{concat}(\x_t,\y)\), significantly improves performance. Building on this, we train a latent diffusion version using a task-specific VAE. However, due to the precision requirements of PDEs, the latent model performs poorly, highlighting the need for pixel-space modeling in scientific applications. Next, we introduce a hierarchical variant of the DiT with 3D neighborhood attention and temporal downsampling inspired by HDiT \cite{crowson2024hourglass}, which further reduces error. Finally, conditioning on the binary mask itself yields the best performance in our setup, indicating that the binary mask resolves ambiguity between masked and unmasked pixels: the input is \(\mathrm{concat}(\x_t,\y,\m)\).

\subsection{Supplementary}
We provide further details on our training hyperparameters, datasets, and model architecture in the supplementary.
Also provided are extended experimental results and analysis for more tasks and PDE problem settings.
We urge readers to review the additional images and videos in the supplementary.


\vspace{-0.5em}
\section{Conclusion} \label{sec:conclusion}
\vspace{-0.2cm}
In this work, we present \ours{}, a unified framework for solving PDEs by reframing them as inpainting problems within a video diffusion model.
We propose a pixel-level, hierarchical video diffusion architecture that leverages a flexible conditioning mechanism, allowing adaptation to arbitrary observation patterns without architectural changes.
Our method significantly improves over the state-of-the-art, and we found that a single model trained on multiple tasks consistently outperforms existing specialized solvers.
As an example of our method’s effectiveness, \ours{} accurately reconstructs 20 frames from just 3\% observations at a single time step (Figure~\ref{fig:inverse}).

In future work, we aim to extend \ours{} to more challenging long-term prediction settings, where uncertainty accumulates over time. We also plan to explore evaluation metrics that better capture stochastic and multimodal behavior beyond pixel-wise error. Finally, we will study 3D generative PDE solvers to handle volumetric and spatiotemporal physical systems.


\section*{Acknowledgment}
This work is supported by the Samsung Global Research Outreach Program.

\bibliographystyle{plain}
\bibliography{references}

\newpage
\appendix

\section{Overview}

This supplementary material provides additional experiments and analyses to support the findings presented in the main paper. In Section \ref{sec:additional_dataset}, we evaluate \ours{} on additional datasets involving both static and dynamic PDEs to highlight its generalization capability. Section \ref{sec:diffusionpde_ext} presents an extension of DiffusionPDE for spatio-temporal PDEs. Section \ref{sec:stochasticity_eval} assesses the model’s robustness to random noise and its inherent stochasticity. In Section \ref{sec:error_rate_vs_obs_rate}, we analyze the relationship between overall error rates and the observation ratio. Section \ref{sec:long_range_pred} investigates the long-range generalization ability of models trained on short-range data. Section \ref{sec:further_ablation} explores the performance of \ours{} under various video settings. Finally, Section \ref{sec:arch_training_detail} includes \ours{}'s architectural and training details.

\subsection{HTML Webpage}
We provide a project page \href{https://videopde.github.io/}{videopde.github.io} with embedded video resources for additional video results showcasing a variety of tasks and PDE families. These visualizations further illustrate the versatility and effectiveness of \ours{} across diverse problem settings. We strongly encourage readers to view the supplemental webpage to fully appreciate the quality of our predictions visually.

\section{Experiments on Additional Dynamic and Static Datasets}\label{sec:additional_dataset}

In this section, we further evaluate \ours{} on additional datasets, including one dynamic and another static equations.

\ours{} can be trivially extended to handle static PDEs by interpreting the coefficient field and the corresponding solution as two distinct time steps within a temporal framework. This perspective enables the application of our dynamic modeling approach to inherently time-independent problems, such as the Helmholtz Equation described below.

\paragraph{Inhomogeneous Helmholtz Equation} Further, we evaluate the static Helmholtz equation as described in DiffusionPDE \cite{huang2024diffusionpdegenerativepdesolvingpartial}.

\begin{equation*}\label{eq:inhomogeneous-helmholtz}
    \begin{aligned}
        \nabla^2\mathbf{u}(\mathbf{c})+\alpha^2\mathbf{u}(\mathbf{c}) & =\mathbf{a}(\mathbf{c}),\quad &  & \mathbf{c}\in\Omega          \\
        \mathbf{u}(\mathbf{c})                                        & =0,                           &  & \mathbf{c}\in\partial\Omega,
    \end{aligned}
\end{equation*}
where $\alpha=1$.

\begin{figure}[H]
    \centering
    \vspace{-2.5em}
    \includegraphics[width=0.99\linewidth]{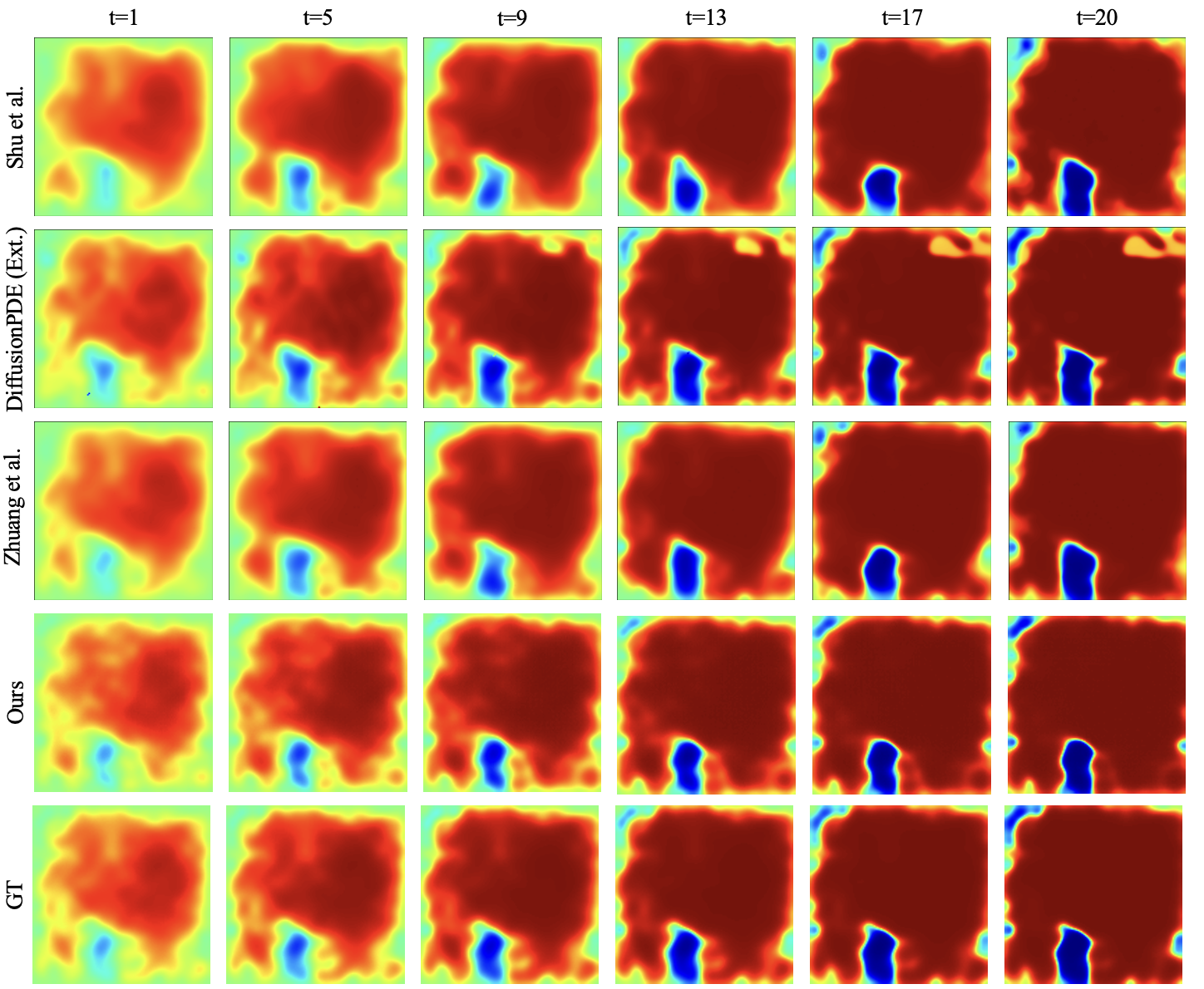}
    \vspace{-0.7em}
    \caption{\textbf{Reconstruction results of ACE dataset with 1\% observation points.} Visualizations of all methods for the continuous partial observation task.}
    \label{fig:ace-sparse-continuous}
\end{figure}

Moreover, to compensate the dynamic PDEs in the main paper, which mostly feature highly dynamic propagations, we test \ours{} on another PDE (Allen-Cahn) that models reaction-diffusion described below. As shown in Fig.~\ref{fig:ace-sparse-continuous}, the slow-moving nature of the solution fields makes them challenging to capture from fixed sensors, resulting in higher overall errors compared to more dynamic datasets in the main paper.

\paragraph{Allen–Cahn Equation} We study the time-dependent Allen-Cahn equation (ACE) task using the dataset prepared by Poseidon \cite{herde2024poseidon}.

\begin{equation*}
    \begin{aligned}
        \partial_t\mathbf{u}(\mathbf{c}, \tau) & =\Delta \mathbf{u}(\mathbf{c}, \tau)-\gamma^2\mathbf{u}(\mathbf{c}, \tau)(\mathbf{u}(\mathbf{c}, \tau)^2-1), \quad &  & \mathbf{c}\in\Omega, \tau\in (0,T] \\
        \mathbf{u}(\mathbf{c}, 0)              & = \mathbf{u}_0(\mathbf{c}),                                                                                        &  & \mathbf{c}\in\Omega
    \end{aligned}
\end{equation*}

where $\gamma=220$ is the reaction rate.

\begin{table}[!htbp]
    \centering
    \vspace{-0.2em}
    \begin{tabular}{
            l 
            rr 
            rr 
        }
        \toprule
                            & \multicolumn{2}{c}{\textbf{Allen–Cahn}} & \multicolumn{2}{c}{\textbf{Helmholtz}}                                            \\
                            & 1\%                                     & 3\%                                    & 1\%                 & 3\%                \\
        \midrule
        DiffusionPDE        & 30.65\%                                 & 7.45\%                                 & \textbf{9.75\%}     & \textbf{6.30\%}    \\
        DiffusionPDE (Ext.) & 29.05\%                                 & \underline{6.93\%}                     & \text{\sffamily X}  & \text{\sffamily X} \\
        Shu et al.          & 30.71\%                                 & 8.10\%                                 & \text{\sffamily X}  & \text{\sffamily X} \\
        Zhuang et al.       & 27.43\%                                 & 7.07\%                                 & \text{\sffamily X}  & \text{\sffamily X} \\
        \midrule
        Ours                & \textbf{23.95\%}                        & \textbf{6.15\%}                        & \underline{10.14\%} & \underline{6.41\%} \\
        Ours (Unified)      & \underline{27.11\%}                     & 7.20\%                                 & 10.72\%             & 6.92\%             \\
        \bottomrule
    \end{tabular}
    \caption{\textbf{Additional continuous partial observation reconstruction.} Quantitative $\ell_2$ relative errors of various methods on the dynamic Allen–Cahn equation and the static Helmholtz equation.}
    \label{tab:supp-continuous}
\end{table}

\begin{table}[!htbp]
    \centering
    \vspace{-1.5em}
    \begin{tabular}{
            l 
            rr 
            rr 
        }
        \toprule
                       & \multicolumn{2}{c}{\textbf{Allen–Cahn}} & \multicolumn{2}{c}{\textbf{Helmholtz}}                                        \\
                       & Forward                                 & Inverse                                & Forward            & Inverse         \\
        \midrule
        DiffusionPDE   & 1.43\%                                  & 3.41\%                                 & 2.3\%              & 4.0\%           \\
        PINO           & 1.39\%                                  & 2.70\%                                 & 4.9\%              & 4.9\%           \\
        \midrule
        Ours           & \textbf{0.55\%}                         & \textbf{1.18\%}                        & \textbf{0.47\%}    & \textbf{3.87\%} \\
        Ours (Unified) & \underline{0.90\%}                      & \underline{1.42\%}                     & \underline{1.07\%} & 4.84\%          \\
        \bottomrule
    \end{tabular}
    \caption{\textbf{Additional forward/inverse full observation.} Average $\ell_2$ relative errors of baseline methods for forward and inverse subtasks on ACE and Helmholtz datasets.}
    \label{tab:supp-fwd-inv-full}
\end{table}

\begin{table}[!htbp]
    \centering
    \vspace{-1.5em}
    \begin{tabular}{
            l 
            rr 
            rr 
        }
        \toprule
                       & \multicolumn{2}{c}{\textbf{Allen–Cahn}} & \multicolumn{2}{c}{\textbf{Helmholtz}}                                            \\
                       & Forward                                 & Inverse                                & Forward            & Inverse             \\
        \midrule
        DiffusionPDE   & 15.29\%                                 & 15.83\%                                & 8.8\%              & 22.6\%              \\
        \midrule
        Ours           & \underline{14.68\%}                     & \underline{13.48\%}                    & \textbf{2.32\%}    & \underline{11.02\%} \\
        Ours (Unified) & \textbf{13.27\%}                        & \textbf{12.64\%}                       & \underline{2.45\%} & \textbf{8.97\%}     \\
        \bottomrule
    \end{tabular}
    \caption{\textbf{Additional forward/inverse 3\% observation.} Average $\ell_2$ relative errors of baseline methods for forward and inverse subtasks on ACE and Helmholtz datasets.}
    \label{tab:supp-fwd-inv-partial}
\end{table}

Similar to the main paper, we assess the performance of various models across three distinct tasks.

\paragraph{Continuous Partial Observation} Based on the settings from the main experiments, we reconstruct complete spatiotemporal and static PDE trajectories using sparse, fixed-point observations. Specifically, we evaluate the average error of the coefficient and solution for static PDEs. Table \ref{tab:supp-continuous} demonstrates that \ours{} consistently produces more accurate reconstructions than all baseline methods on the dynamic PDE task, as shown in Figure \ref{fig:ace-sparse-continuous} as well, reinforcing the findings reported in the main paper. Furthermore, \ours{} achieves performance comparable to state-of-the-art methods on static PDE reconstruction when there are observations on both coefficients and solutions.

\paragraph{Forward/Inverse Full Observation}
Similarly, we evaluate the results of both forward and inverse problems for these additional datasets. In the context of static PDEs, the forward problem involves predicting the solution space given the coefficient space, whereas the inverse problem entails inferring the coefficient space from the observed solution space. Table \ref{tab:supp-fwd-inv-full} shows that \ours{} outperforms all baselines on fully observed forward and inverse problems across all PDE families, including both dynamic and static cases.

\begin{figure}[h]
    \centering
    \includegraphics[width=0.8\linewidth]{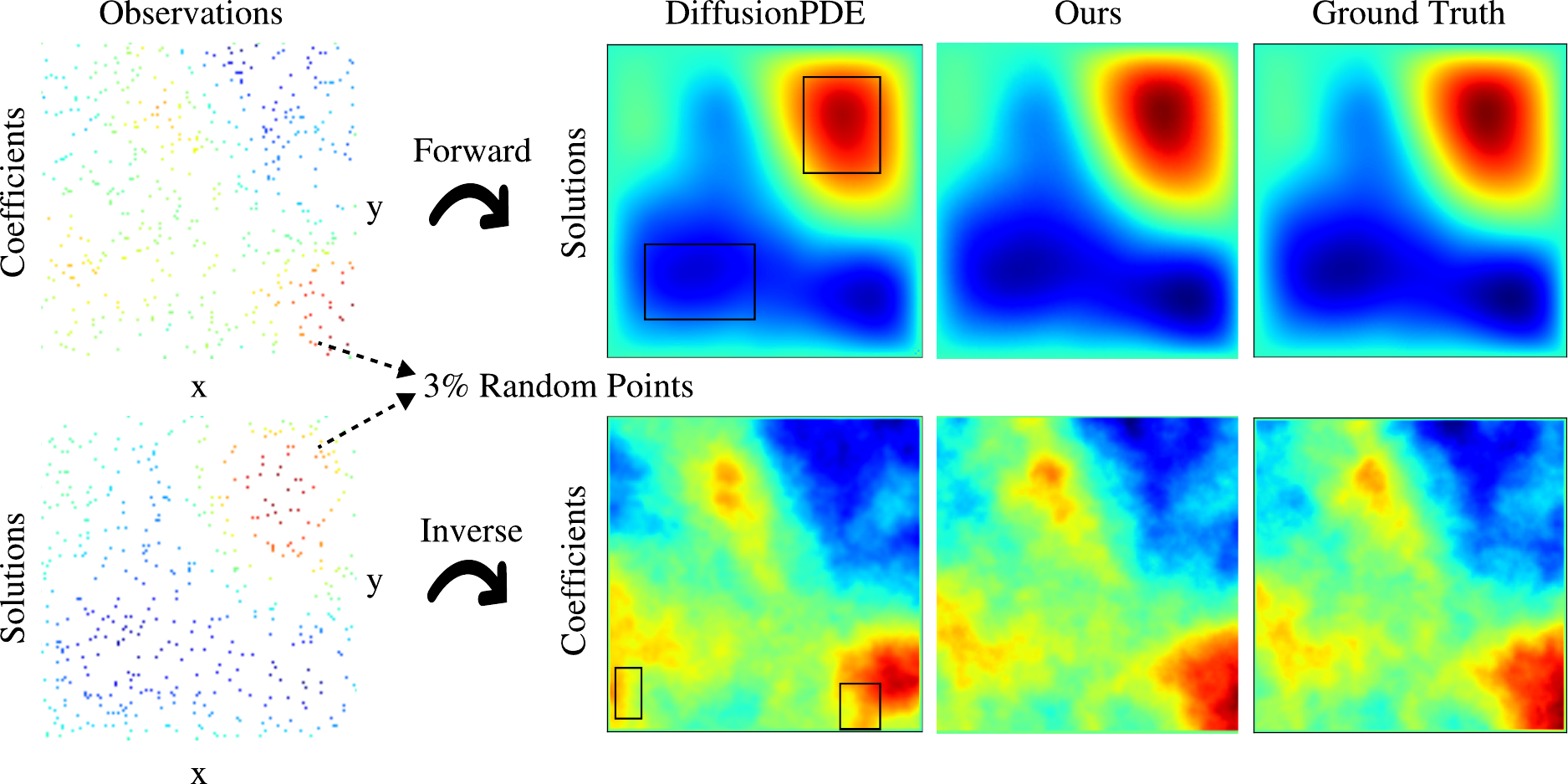}
    \caption{\textbf{Comparison of partially-observed forward/inverse predictions on Helmholtz equation.} Results of different methods for forward/inverse from partial observations (3\%) are compared along with the ground truth fields. Notable errors produced by DiffusionPDE are highlighted with black squares for emphasis.
    }
    \label{fig:supp:helmholtz-fwd-inv-sparse}
\end{figure}

\paragraph{Forward/Inverse Partial Observation} We further investigate the forward and inverse problems using 3\% observation points on the respective side. In Table \ref{tab:supp-fwd-inv-partial}, we show that \ours{} has a lower error regarding both forward and inverse processes compared with DiffusionPDE. \ours{} tends to generate more accurate results, as present in Figure \ref{fig:supp:helmholtz-fwd-inv-sparse}.

\section{DiffusionPDE Extension}\label{sec:diffusionpde_ext}

\begin{figure}[h]
    \centering
    \includegraphics[width=0.8\linewidth]{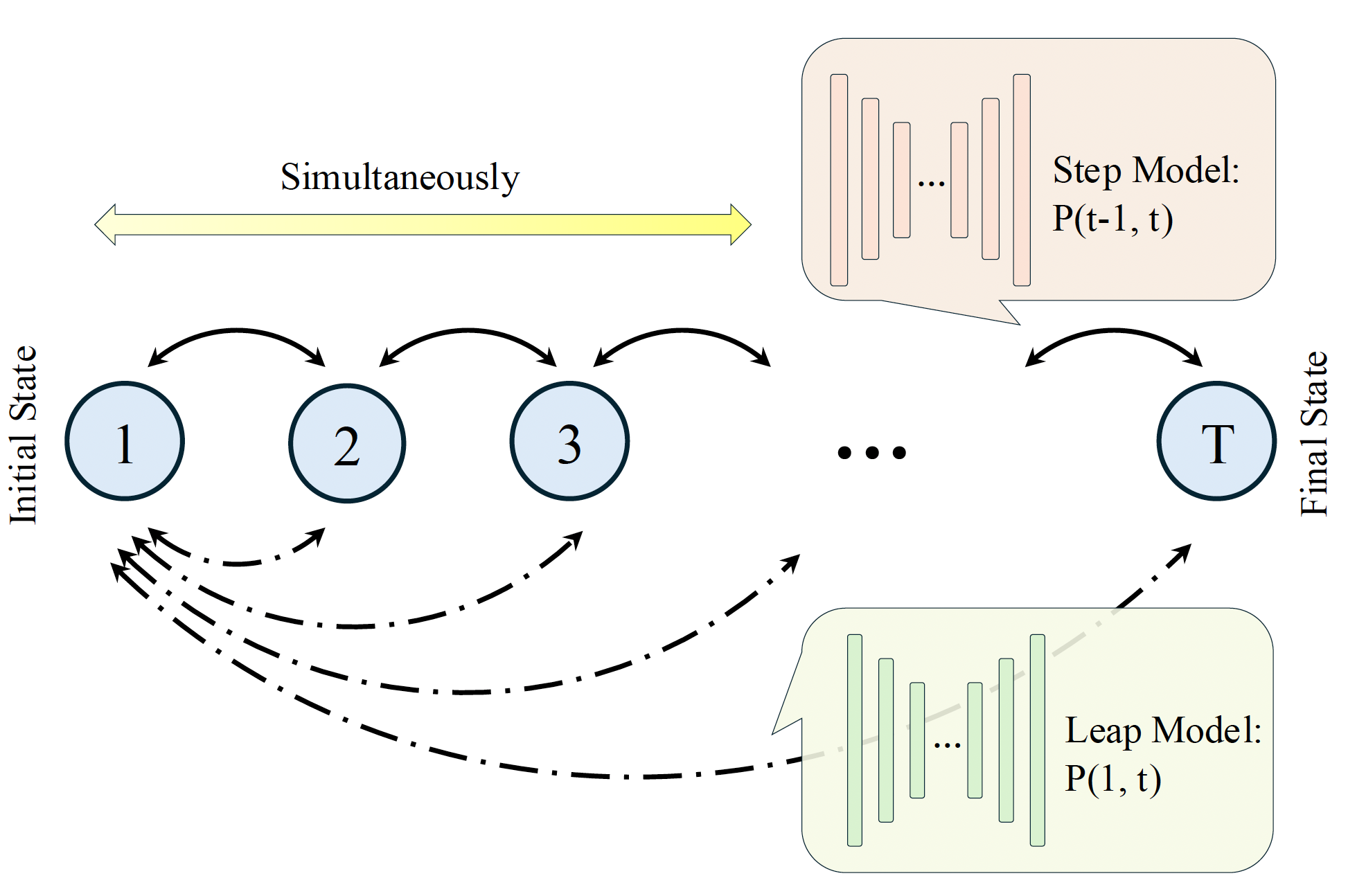}
    \caption{\textbf{Extension framework of DiffusionPDE.} The step model for distributing two adjacent frames and the leap model for distributing the initial state alongside one random timestep jointly denoise all time steps in a simultaneous manner.
    }
    \label{fig:diffusionpde_ext}
\end{figure}

In this section, we explain how we extend the DiffusionPDE framework \cite{huang2024diffusionpdegenerativepdesolvingpartial} to address dense temporal predictions by introducing a two-model architecture tailored for the continuous partial observation setting, as illustrated in Figure~\ref{fig:diffusionpde_ext}. Specifically, we employ a step model and a leap model to learn the joint distributions over adjacent timesteps, $P(t-1, t)$, and between the initial state and an arbitrary timestep, $P(1, t)$, respectively, where $t=2, 3, \dots, T$ is the index of the timestep. Both models are explicitly conditioned on the timestep $t$. The final score for $P(t)$ is computed by averaging the denoised estimates from the two pre-trained models, as demonstrated in Algorithm \ref{alg:diffusionpde-ext}, thereby integrating both local and long-range temporal information.

\begin{algorithm}[!tbp]
	\caption{DiffusionPDE (Ext.) Joint Denoising Algorithm.}
	\label{alg:diffusionpde-ext}
	\begin{algorithmic}[1]
		\State \textbf{input} StepDenoiser $D_S(t)$, LeapDenoiser $D_L(t)$, VideoLength $T$, Timestep $t=1, 2, \dots, T$, TotalIterationCount $N$
            \For{$i \in \{0, \dots, N-1\}$}
                \For{$j \in \{2, \dots, T\}$}
                    \State $x^{j-1}_S, x^j_S \leftarrow D_S(j)$ \mycomment Denoise the step model at timestep $j$
                    \State $x^{1}_L, x^{j}_L \leftarrow D_L(j)$ \mycomment Denoise the leap model at timestep $j$
                    \State $x_j \leftarrow (x_S^j+x_L^j)/2$ \mycomment Average the scores
                    \State $x_{j-1} \leftarrow (x_S^{j-1}+x_L^{j-1})/2$
                    \State \{\dots Further guided sampling steps\dots\}
                \EndFor                
            \EndFor
            \State \textbf{return} $x_1, \dots, x_T$
    \end{algorithmic}  
\end{algorithm}

This method can also be extended to both forward and inverse problems where the only observation is on the initial or final frame. However, rather than denoising all frames simultaneously, these tasks required an autoregressive approach to operate reasonably, which significantly increases computational cost. As a result, we do not include forward/inverse predictions with this extended model.

\section{Multi-modality and Robustness Evaluation}\label{sec:stochasticity_eval}

\begin{figure}[H]
    \centering
    \includegraphics[width=0.99\linewidth]{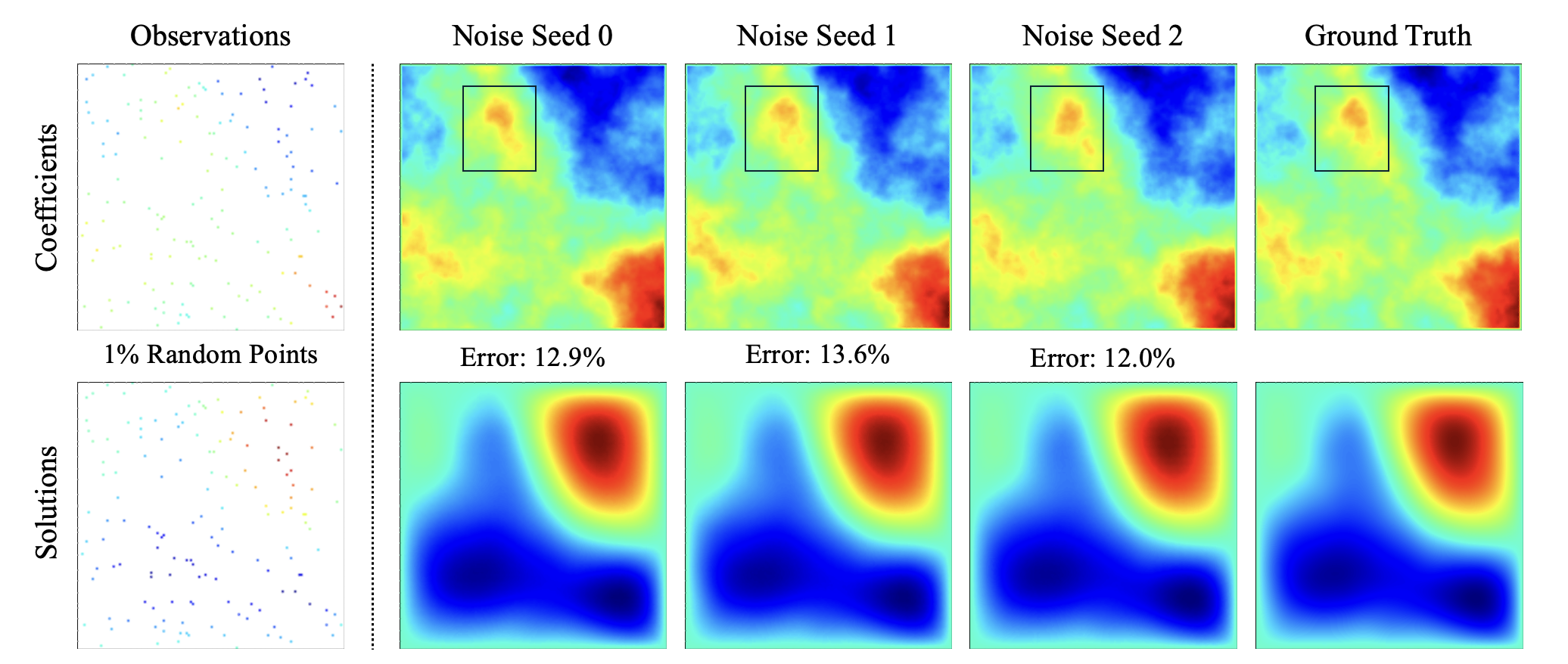}
    \caption{\textbf{Helmholtz reconstructions with different initial noise seeds.} Results and corresponding $\ell_2$ relative errors obtained under identical 1\% observation on both the coefficient and solution domains. Significant differences are emphasized with black squares.}
    \label{fig:helmholtz_different_noise_seed}
\end{figure}

To study the stochasticity and robustness of \ours{}, we present reconstructions generated using different initial noise realizations of the diffusion model in Figure~\ref{fig:helmholtz_different_noise_seed}. Given a fixed observation mask, \ours{} produces diverse reconstructions, particularly in regions lacking observations, which aligns with the physics property. Despite this variability, the model demonstrates strong robustness, as the reconstructed solutions and corresponding error metrics remain consistent across different noise seeds.

\section{Error Rate versus Observation Rate}\label{sec:error_rate_vs_obs_rate}

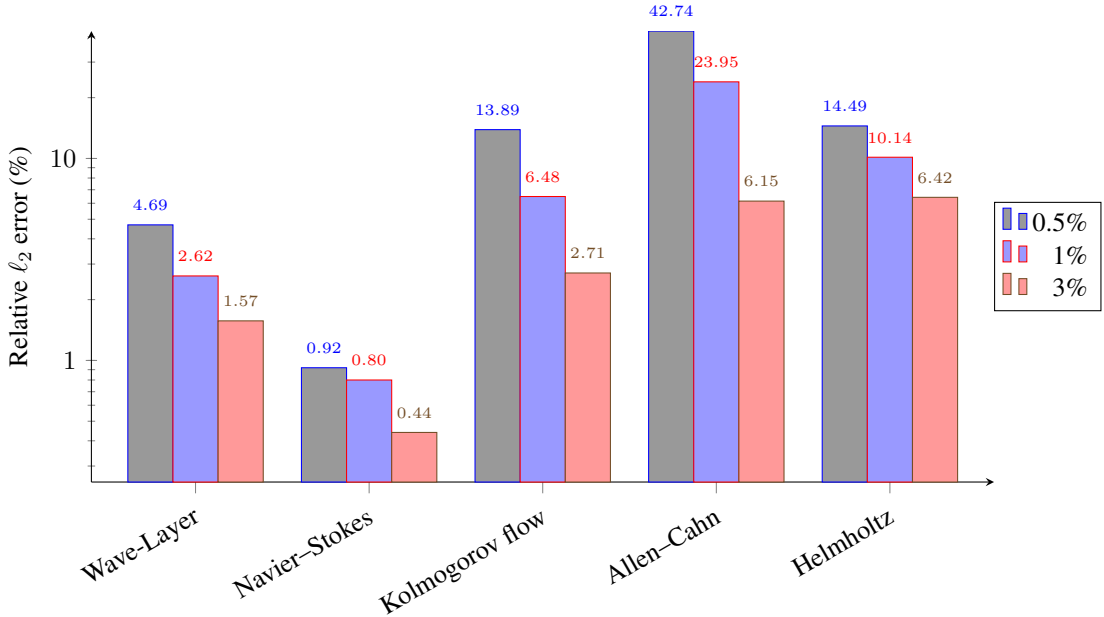
\begin{figure}[ht]
    \makebox[\textwidth][c]{
        \begin{tikzpicture}
  \begin{axis}[
    height=6cm,
    width=12cm,
    scale only axis,
    axis x line=bottom,
    axis y line=left,
    ybar=0pt,
    ymode=log,
    ymin=0.25,
    log origin=infty,
    bar width=0.6cm,
    enlarge x limits=0.15,
    ylabel={Relative \(\ell_2\) error (\%)},
    symbolic x coords={WL,NS,KF,ACE,Helmholtz},
    xtick=data,
    xticklabels={Wave-Layer,Navier–Stokes,Kolmogorov flow,Allen–Cahn,Helmholtz},
    xticklabel style={
        align=center,
        rotate=30,
        anchor=north east,
    },
    log ticks with fixed point,
    minor y tick num=1,
    point meta=rawy,
    nodes near coords,
    every node near coord/.append style={
      font=\tiny,
      yshift=2pt,
      /pgf/number format/.cd,
        fixed,
        precision=2,
        zerofill
    },
    legend style={
      at={(rel axis cs:1,0.5)},
      anchor=west,
      legend columns=1,
    },
    legend cell align=right,
  ]
    \addplot+[fill=black!40] coordinates {
      (WL,4.69) (NS,0.92) (KF,13.89) (ACE,42.74) (Helmholtz,14.49)
    };
    \addplot+[fill=blue!40] coordinates {
      (WL,2.62) (NS,0.80) (KF,6.48) (ACE,23.95) (Helmholtz,10.14)
    };
    \addplot+[fill=red!40] coordinates {
      (WL,1.57) (NS,0.44) (KF,2.71) (ACE,6.15) (Helmholtz,6.42)
    };
    \legend{0.5\%,1\%,3\%}
  \end{axis}
\end{tikzpicture}
    }
    \caption{\textbf{Continuous partial observation comparison.} We report relative \(\ell_2\) errors for \ours{} at \(0.5\%\), \(1\%\), and \(3\%\) observation rates.}
    \label{fig:supp:obs-vs-error}
\end{figure}

We present a chart illustrating the relationship between the error rate and observation rate for the continuous partial observation task, as shown in Figure \ref{fig:supp:obs-vs-error}. The results demonstrate that our methods are can achieve relative \(\ell_2\) errors below 10\% using only 3\% of observation points across all PDE families. For PDEs with high spatial information, e.g., KF, or diffusive behavior, e.g., Allen-Cahn, the performance degrades relatively steeply as the observation ratio decreases.

\section{Long Range Prediction}\label{sec:long_range_pred}

We further assess the performance of all models over an extended time horizon. Specifically, we employ autoregressive prediction to generate 100 future frames of the KF dataset using a 20-frame forward model. The sole input provided is the complete observation of the initial frame.
Each 20-frame window is conditioned on the final frame of the previous window.
The windows are concatenated to produce the final long trajectory.

\begin{figure}[h]
    \centering
    \begin{subfigure}[b]{0.48\textwidth}
        \centering
        \includegraphics[width=\textwidth]{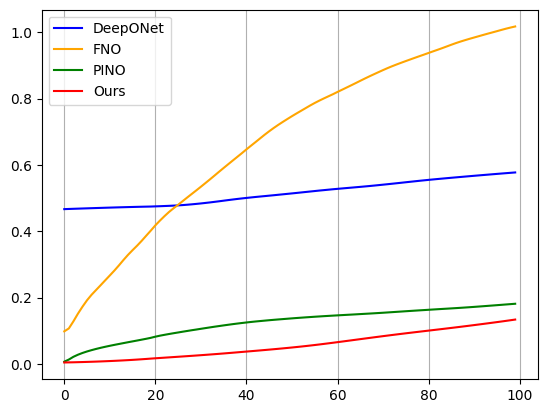}
        \caption{Average errors of the first $n$ frames.}
        \label{fig:long-range-average}
    \end{subfigure}
    \hfill
    \begin{subfigure}[b]{0.48\textwidth}
        \centering
        \includegraphics[width=\textwidth]{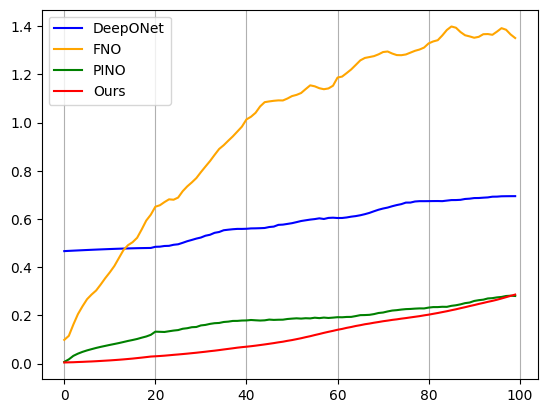}
        \caption{Errors of each single frame.}
        \label{fig:long-range-per-frame}
    \end{subfigure}
    \caption{\textbf{Long-range prediction results.} The $\ell_2$ relative errors of all methods plotted against the total number of frames.}
    \label{fig:long-range-all}
\end{figure}

In Fig. \ref{fig:long-range-all}, we report the overall average errors as well as the single-frame errors for all evaluated models. The results indicate that \ours{} achieves better predictive performance compared to baseline approaches. Nonetheless, a notable accumulation of error is observed for all methods, which may be attributed to the inherent uncertainty and difficulty of predicting long-range future and the absence of explicit PDE constraints for our method. Addressing this limitation remains an important direction for future research.
We refer readers to the attached HTML file to view the videos of the 100-frame predictions.

\section{Further Ablations}\label{sec:further_ablation}

We revisit the Wave-Layer dataset to conduct detailed ablations that study the sensitivity and impact of individual parameter choices on the model's overall performance. Specifically, we examine the influence of varying the number of frames, timestep sizes, and frames per second (FPS) within fixed physics intervals.

\paragraph{Number of frames}

We first investigate how the number of frames affects our model's performance. From the available 20 frames in our raw data, we select subsets consisting of 5, 10, and 20 consecutive frames. All other experimental settings, including model configuration, remain consistent. The results summarized in Table \ref{tab:num_frames} demonstrate that increasing the number of frames leads to improved performance, highlighting the effectiveness of our method in capturing temporal dependencies and longer temporal PDE dynamics. Additionally, since the error is averaged over all time frames, it suggests that utilizing longer temporal sequences improves the model's capability to accurately inpaint initial frames by providing richer temporal context.

\noindent
\begin{minipage}{\linewidth}
    \centering
    \begin{tabular}{lccc}
        \toprule
             & 5 frames & 10 frames & 20 frames \\
        \midrule
        Ours & 0.62\%   & 0.52\%    & 0.49\%    \\
        \bottomrule
    \end{tabular}
    \captionof{table}{\textbf{Number of frames}.}
    \label{tab:num_frames}
\end{minipage}

\paragraph{Timestep size}

Next, we explore how variations in the timestep size between frames impact the model's performance. Specifically, we fix the total number of frames to 5 and alter the interval between frames to step sizes of 1, 2, and 4, corresponding respectively to physical simulation intervals of 0.05s, 0.1s, and 0.2s within the total 1-second simulation duration. Table \ref{tab:timestep} summarizes the observed model errors across these different timestep sizes: smaller timesteps lead to reduced errors as expected.

\noindent
\begin{minipage}{\linewidth}
    \centering
    \begin{tabular}{lccc}
        \toprule
             & step=1 & step=2 & step=4 \\
        \midrule
        Ours & 0.49\% & 0.57\% & 0.94\% \\
        \bottomrule
    \end{tabular}
    \captionof{table}{\textbf{Timestep size}.}
    \label{tab:timestep}
\end{minipage}

\paragraph{Frames per second (FPS)}

Finally, we study the effect of FPS by varying both the number of frames and the timestep sizes to represent the same total physics duration of 1 second. Specifically, we compare scenarios using 5 frames with a timestep of 4, 10 frames with a timestep of 2, and 20 frames with a timestep of 1. As shown in Table \ref{tab:equal_time}, increasing the FPS improves the model's accuracy, showing the importance of temporal granularity in accurately modeling PDE dynamics.

\noindent
\begin{minipage}{\linewidth}
    \centering
    \begin{tabular}{lccc}
        \toprule
             & FPS = 5 & FPS = 10 & FPS = 20 \\
        \midrule
        Ours & 0.94\%  & 0.60\%   & 0.49\%   \\
        \bottomrule
    \end{tabular}
    \captionof{table}{\textbf{Frames per second}.}
    \label{tab:equal_time}
\end{minipage}

\section{Architecture and Training Details}\label{sec:arch_training_detail}
We report our architectural and training hyperparameters in Table \ref{tab:hyperparameters}.

\begin{table}[ht]
    \centering
    \begin{tabular}{
            l
            c
            c
        }
        \toprule
        Hyperparameter                                   & Ours                & Ours (Unified)      \\
        \midrule
        Parameters                                       & 118M                & 74M                 \\
        Training steps                                   & 50k                 & 100k                \\
        Batch size                                       & 48                  & 64                  \\
        GPUs                                             & 2 \(\times\) L40S   & 4 \(\times\) L40S   \\
        Mixed Precision                                  & bfloat16            & bfloat16            \\
        \midrule
        Patch Size (\(T\times H\times W\))               & [2, 4, 4]           & [2, 4, 4]           \\
        Neighborhood Attention Levels                    & 1                   & 1                   \\
        Global Attention Level                           & 1                   & 1                   \\
        Neighborhood Attention Depth                     & 2                   & 2                   \\
        Global Attention Depth                           & 11                  & 6                   \\
        Feature Dimensions                               & [384, 768]          & [384, 768]          \\
        Attention Head Dimension                         & 64                  & 64                  \\
        Neighborhood Kernel Size (\(T\times H\times W\)) & [2, 7, 7]           & [2, 4, 4]           \\
        Mapping Depth                                    & 1                   & 1                   \\
        Mapping Width                                    & 768                 & 768                 \\
        Dropout                                          & 0                   & 0                   \\
        \midrule
        Optimizer                                        & AdamW               & AdamW               \\
        Learning Rate                                    & \(5\times 10^{-4}\) & \(5\times 10^{-4}\) \\
        \([\beta_1,\beta_2]\)                            & \([0.9, 0.95]\)     & \([0.9, 0.95]\)     \\
        Epsilon                                          & \(1\times 10^{-8}\) & \(1\times 10^{-8}\) \\
        Weight Decay                                     & \(1\times10^{-2}\)  & \(1\times10^{-2}\)  \\
        \bottomrule
    \end{tabular}
    \caption{\textbf{Training and inference hyperparameters.}}
    \label{tab:hyperparameters}
\end{table}

\section{Additional Baseline Results}

\begin{figure}[H]
    \centering
    \includegraphics[width=0.6\linewidth]{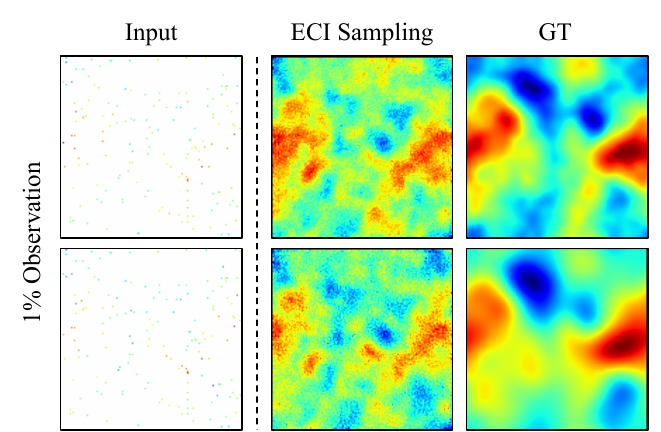}
    \caption{\textbf{ECI sampling results.} Reconstruction results with 1\% observation points of the Navier-Stokes equation.}
    \label{fig:eci}
\end{figure}

We further evaluate the performance of the ECI sampling framework \cite{cheng2025gradientfree}, which is an effective method for handling hard-constrained systems, particularly under various boundary conditions. However, as illustrated in Figure \ref{fig:eci}, the model encounters difficulties when working with sparse observations. This issue may arise from inconsistent guidance within the domain, leading to challenges in global reconstruction, especially with extremely sparse input points. Furthermore, since the data does not have strong boundary conditions, this method has significant limitations for the tasks we discuss.

\end{document}